\documentclass{article}

\PassOptionsToPackage{numbers, compress}{natbib}
\usepackage[preprint]{neurips_2026}

\usepackage[utf8]{inputenc} 
\usepackage[T1]{fontenc}    
\usepackage{hyperref}       
\usepackage{url}            
\usepackage{booktabs}       
\usepackage{amsfonts}       
\usepackage{nicefrac}       
\usepackage{microtype}      
\usepackage{xcolor}         
\usepackage{capt-of}
\usepackage{caption}

\usepackage{enumitem}
\usepackage{xspace}
\usepackage{multirow}
\usepackage[table]{xcolor}
\usepackage{colortbl}
\usepackage{siunitx}
\usepackage{float}
\usepackage{url}
\usepackage{hyperref}

\usepackage{amsmath}
\usepackage{amssymb}
\usepackage{mathtools}
\usepackage{amsthm}
\usepackage{wrapfig}
\usepackage{algorithm}
\usepackage{algpseudocode}

\usepackage[capitalize,noabbrev]{cleveref}

\theoremstyle{plain}

\theoremstyle{definition}

\theoremstyle{remark}

\newcommand{\ourmethod}{\textsc{MemCollab}\xspace}

\title{MemCollab: Cross-Model Memory Collaboration via Contrastive Trajectory Distillation}

\author{%
  Yurui Chang$^{1}$ \\
  \texttt{yuruic@psu.edu} \\
  \And
  Yiran Wu$^{1}$ \\
  \texttt{yiran.wu@psu.edu} \\
  \And
  Qingyun Wu$^{2}$ \\
  \texttt{qingyun@ag2.ai} \\
  \And
  Lu Lin$^{1}$ \\
  \texttt{lulin@psu.edu} \\
  \AND
  $^{1}$Pennsylvania State University \quad
  $^{2}$AG2AI
}

\begin{document}

\maketitle

\begin{abstract}
LLM agents increasingly rely on memory mechanisms to reuse knowledge from past problem-solving experiences. However, existing methods typically construct memory for a single agent and reuse it with the same underlying model, tightly coupling stored knowledge to model-specific reasoning styles. In heterogeneous deployments, where agents may be instantiated with backbone models of different sizes, architectures, or specializations, this raises a key question: \emph{can a single memory system be shared across agents with different backbone models?}
We find that naive cross-model memory transfer can degrade performance, because stored memories often entangle task-relevant knowledge with model-specific biases. To address this challenge, we propose \ourmethod, a collaborative memory framework that builds shared cross-model memory by contrasting reasoning trajectories generated by different model-based agents on the same task. Through this contrastive process, \ourmethod distills abstract reasoning constraints that capture shared task-level invariants while suppressing model-specific artifacts. We further introduce a task-aware retrieval mechanism that conditions memory access on task category, ensuring that only relevant constraints are retrieved at inference time.
Experiments on mathematical reasoning and code generation benchmarks show that \ourmethod consistently improves both accuracy and inference-time efficiency across diverse agents, including settings with different model families. These results demonstrate that collaboratively constructed cross-model memory can serve as a shared reasoning resource for heterogeneous LLM-based agents.

\end{abstract}

\section{Introduction}

Large language model (LLM)-based agents are increasingly used to solve complex tasks involving multi-step reasoning, tool use, and interaction with external environments. Despite their strong capabilities, they remain limited by the lack of persistent episodic knowledge: information from prior problem-solving trajectories is not naturally retained, and inference is constrained by a fixed context window~\citep{laban2025llms, kang2025acon}. As a result, agents may repeatedly explore similar strategies, encounter recurring errors, and fail to reuse useful tool feedback~\citep{pink2025position, anthropic2025context}. Memory mechanisms address these limitations by storing and retrieving reusable information from past experiences for future inference~\citep{zhong2024memorybank, xu2025mem}.

Recent studies show that memory-augmented LLM agents can improve reasoning performance. For example, BoT~\citep{yang2024buffer} distills reasoning templates from past experiences and retrieves them during inference. However, most existing approaches construct memory in a model- or agent-specific manner, deriving it from an individual agent's own reasoning traces and reusing it with the same agent. This tightly couples stored knowledge to the reasoning style and biases of the agent's underlying backbone model. Yet modern LLM deployments increasingly involve heterogeneous agents instantiated with backbone models of different sizes, architectures, or specializations, operating through routing, orchestration, or collaboration. This motivates a natural question: \textit{can a single memory system be shared across agents with different backbone models?}

\begin{wrapfigure}{r}{0.50\textwidth}
    \centering
    \includegraphics[width=0.48\textwidth]{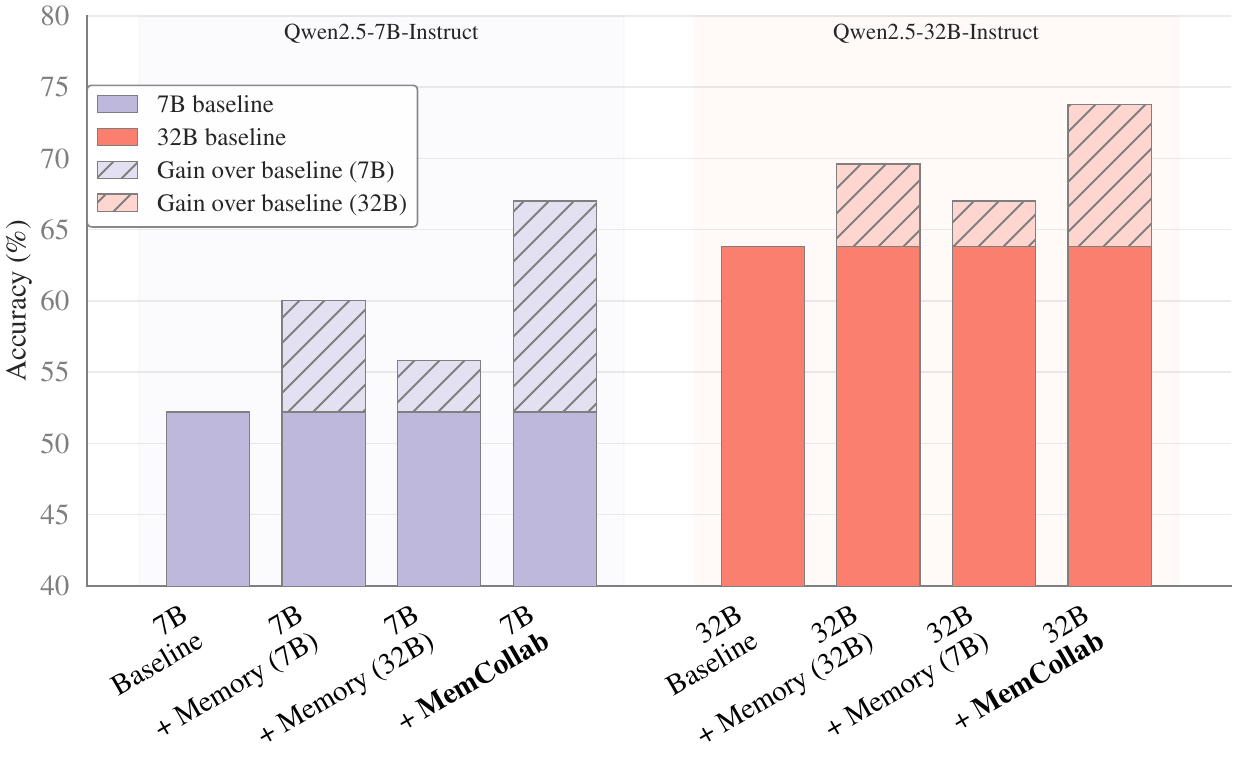}
    \caption{Accuracy on MATH500 dataset with memory from different resources.}
    \vspace{-1em}
    \label{fig:math500-boxplot}
\end{wrapfigure}

A naive approach is to directly transfer memory built from one model-based agent to another. In Figure~\ref{fig:math500-boxplot}, we study how external memory affects agent performance. We compare the no-memory baseline to variants that use memory built from the same agent's trajectories and memory built from trajectories generated by another model-based agent. The results indicate that directly reusing memory across backbone models can perform worse than using memory distilled from the target agent's own trajectories. This suggests that memory often reflects the originating model's preferences and reasoning style, such as preferred solving strategies and habitual heuristics, rather than objective and transferable knowledge. This motivates studying whether we can construct more transferable memory by leveraging and contrasting trajectories from agents instantiated with different backbone models, extracting shared and reusable guidance while reducing model-specific bias.

In this paper, we propose \ourmethod{}, a memory framework designed to produce \emph{shared cross-model memory} for agents instantiated with heterogeneous backbone models. Such memory can be reused across agents with different backbone models while \textbf{outperforming memory distilled from a single agent's own trajectories} (Figure~\ref{fig:math500-boxplot}). 
\ourmethod{} extracts reusable reasoning guidance and tool-usage instructions by identifying shared invariances and discrepancies across trajectories generated by different model-based agents, instead of encoding model-specific preferences. This contrastive approach filters out model-specific biases while preserving transferable reasoning constraints and error-forbidden patterns. To ensure that memory entries capturing task-relevant reasoning procedures and error patterns are retrieved for a specific task, we design a task-aware retrieval mechanism that conditions memory retrieval on the task category. We summarize our contributions below:

\begin{itemize}[leftmargin=*]
    \setlength\itemsep{-0.01em}
    \item We identify and study the problem of constructing shared memory for multiple LLM-based agents with different backbone models, a setting increasingly common in modern deployments but underexplored in prior agent-memory work.
    
    \item We introduce a cross-model memory collaboration paradigm that derives memory through contrastive analysis of reasoning traces from different agents on the same task. Rather than storing raw experiences, it distills transferable reasoning strategies and cross-model failure patterns into reusable guidance.

    \item Rigorous evaluations on mathematical reasoning and code generation benchmarks show that \ourmethod{} improves agents built on both weaker and stronger backbones, while also enhancing inference-time efficiency through compact, low-noise reasoning guidance.
\end{itemize}
\section{Methodology}
\label{sec:method}
\begin{figure*}[!ht]
    \centering
    \includegraphics[width=.8\textwidth]{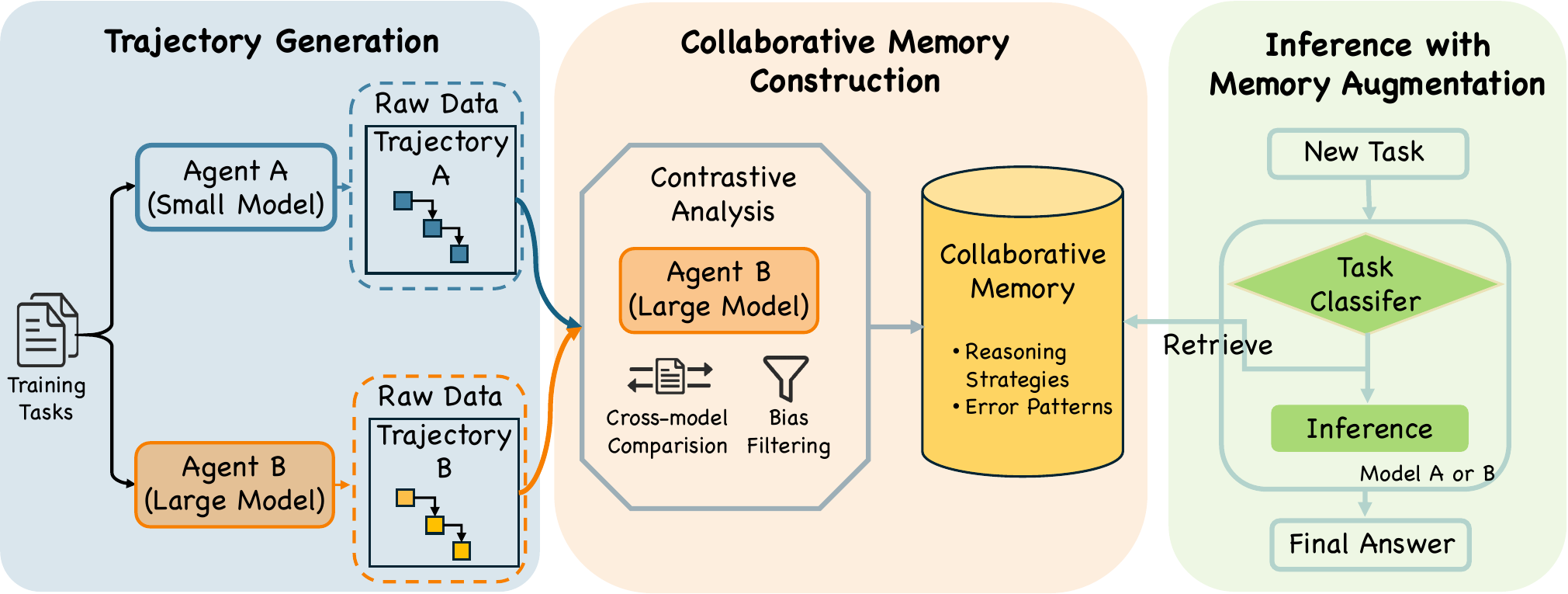}
    \caption{Framework of \ourmethod. \textit{Left}: Using different agents to generate trajectory pairs for the same task. \textit{Middle}: Contrasting the trajectory pairs to construct the memory. \textit{Right}: Augmenting inference-time generation with retrieved memory.}
    \label{fig:framework}
\end{figure*}

We propose \ourmethod, a memory collaboration framework that improves the agent's reasoning by inducing contrast-derived problem-solving strategies from heterogeneous agent trajectories. The core idea is to distill reusable high-level reasoning strategies and error-forbidden constraints that distinguish successful reasoning from failures, while filtering out agent-specific biases. \ourmethod consists of two stages: (1) \textbf{contrast-derived memory construction}, which extracts abstract reasoning constraints from paired agent trajectories; and (2) \textbf{task-aware memory retrieval and inference}, which conditions agent reasoning on relevant memory entries at test time.
An overview of the framework is shown in Figure~\ref{fig:framework}.

\subsection{{Problem Setup and Notation}}
Let $\mathcal{A}=\{A_i\}_{i=1}^N$ denote a set of agents with heterogeneous
reasoning capacities, such as different model sizes, architectures, or training procedures. Each agent $A_i$ is instantiated with a base backbone LLM $f_i$ and produces a multi-step reasoning trajectory when solving a task. Given a task instance $x$, each agent independently generates a reasoning trajectory
\begin{equation}
\tau_i^{(x)} = (r_{i,1}, r_{i,2}, \dots, r_{i,T_i}),
\end{equation}
where each element $r_{i,t}$ represents an intermediate reasoning step from
agent $A_i$, which may incorporate natural language reasoning, code, and code
execution feedback. We assume access to a training set of tasks $\mathcal{D}$,
each with a ground-truth solution or executable verifier. Our goal is to
construct a shared memory bank $\mathcal{M}$ that can be retrieved and reused by multiple agents at inference time.

\subsection{Contrast-Derived Memory Construction}

Algorithm~\ref{alg:construct} summarizes the detailed procedure for memory bank construction, with major steps explained as below.

\textbf{Trajectory Pairing and Preference Selection.}
For each task $x\in\mathcal{D}$, we collect $N$ trajectories
$\{\tau_i\}_{i=1}^N$ by executing multiple agents or model variants on the same input. Let $i^\star$ denote the strongest model, and let $\mathcal{I}(\cdot)$ be an indicator function that evaluates whether a trajectory is correct with respect to the task ground truth, such as by checking the final answer or execution result. We select the preferred trajectory as either a correctly executed one or the one of the strongest model, denoted as:
\begin{equation}
\tau^+ =
\begin{cases}
\tau \sim \mathrm{Unif}\!\left(\{\tau_i \mid \mathcal{I}(\tau_i)=1\}\right),
& \text{if } \mathcal{I}(\tau_{i^\star})=0
\text{ and } \exists i \neq i^\star,\ \mathcal{I}(\tau_i)=1,\\
\tau_{i^\star},
& \text{otherwise.}
\end{cases}
\end{equation}
The remaining trajectories are treated as unpreferred trajectories:
\begin{equation}
    \mathcal{T}^-=\{\tau_i\}_{i=1}^N\setminus\{\tau^+\}.
\end{equation}
This formulation generalizes pairwise trajectory selection to the multi-trajectory setting. It allows the preferred trajectory to originate from any model that successfully solves the task, rather than always from the strongest model, making the preference signal robust to strongest-model failures. The unpreferred set provides contrastive references for extracting reasoning errors and model-specific failure patterns.

\textbf{Contrastive Trajectory Analysis.}
Rather than performing step-by-step comparison, \ourmethod summarizes trajectory differences at the structural reasoning level, including high-level solution algorithms, planning decisions, and the interpretation of tool feedback. Given the preferred trajectory $\tau^+$ and the set of unpreferred trajectories $\mathcal{T}^-$, we contrast $\tau^+$ with each
$\tau^- \in \mathcal{T}^-$  independently and summarize their differences into structured guidance. The summarization is performed by the backbone model of the strongest agent, $f_{i^\star}$, or a designated summarizer model; see Appendix~\ref{appendix:prompt} for the detailed prompt.

For each contrastive pair $(\tau^-, \tau^+)$, we distill two types of information: (1) \textbf{violation patterns} $v_k$, which characterize systematic reasoning failures observed in $\tau^-$, such as premature computation, incorrect symbolic mapping, or misinterpretation of tool feedback; and (2) \textbf{reasoning invariants} $i_k$, which are preserved in $\tau^+$ but absent or violated in $\tau^-$, such as structural principles, planning constraints, or calibration rules required to avoid the corresponding failure.  Analyzing over each contrastive pair will result in a discrepancy set:
\begin{equation}
\Delta(\tau^-, \tau^+) = \{(v_k, i_k)\}_{k=1}^{K},
\end{equation}

\textbf{Memory Bank Construction.} 
Given the extracted discrepancy $(v_k, i_k)$, each contrastive pair is converted into a normative reasoning constraint: 
\begin{equation} 
m_k = (\text{enforce } i_k;\ \text{avoid } v_k), 
\end{equation} 
Each extracted entry is further tagged with a model-identity label, indicating the agent or model that produced the trajectories. These model-identity labels do not create separate memory banks; rather, they enable personalized retrieval from a shared collaboratively constructed memory bank, reducing interference while preserving cross-model abstraction. 

Specifically, $\ell^+$ denotes the identity of the agent that generates the preferred trajectory $\tau^+$ for task $x$, and $\ell^-$ denotes the identity of the agent that generates the unpreferred trajectory $\tau^-$. For the multi-trajectory setting, the extracted memories for task $x$ are defined as: 
\begin{equation} 
\mathcal{M}^{(x)} = \bigcup_{\tau_{l^-} \in \mathcal{T}^-} 
\left\{ \left(\{m_k\}_{k=1}^K, \ell^+, \ell^-\right) \right\}. 
\end{equation} 
\begin{wrapfigure}{r}{0.48\textwidth}
\vspace{-1em}
\centering
\footnotesize

\begin{minipage}{0.48\textwidth}
\hrule
\vspace{0.5em}

\refstepcounter{algorithm}
\label{alg:construct}
\noindent\textbf{Algorithm~\thealgorithm: Multi-model Memory Construction}

\vspace{0.5em}
\hrule
\vspace{0.5em}

\begin{flushleft}
\vspace{-4pt}
\textbf{Input:} Agents $\mathcal{A}=\{A_i\}_{i=1}^{N}$; tasks $\mathcal{D}$;
summarizer $f_{i^\star}$; strongest index $i^\star$; evaluator $\mathcal{I}(\cdot)$; max entries $K$.\\
\textbf{Output:} Labeled memory bank $\mathcal{M}$.
\end{flushleft}

\vspace{-0.8em}

\begin{algorithmic}[1]
\State $\mathcal{M}\leftarrow \emptyset$

\For{each task $x\in\mathcal{D}$}

    \Statex \hspace{\algorithmicindent}\(\triangleright\) collect multi-agent trajectories
    \For{$i=1,\ldots,N$}
        \State $\tau_i\leftarrow A_i(x)$
    \EndFor

    \Statex \hspace{\algorithmicindent}\(\triangleright\) select preferred trajectory
    \If{$\mathcal{I}(\tau_{i^\star})=0$ and $\exists i\neq i^\star,\ \mathcal{I}(\tau_i)=1$}
        \State $l^+\sim \mathrm{Unif}\!\left(\{i\mid \mathcal{I}(\tau_i)=1\}\right)$
    \Else
        \State $l^+\leftarrow i^\star$
    \EndIf
    \State $\tau^+\leftarrow \tau_{l^+}$; $\mathcal{T}^-\leftarrow \{\tau_i\}_{i=1}^{N}\setminus\{\tau^+\}$

    \Statex \hspace{\algorithmicindent}\(\triangleright\) construct preference-pair labeled memory
    \For{each $\tau_j^-\in\mathcal{T}^-$}
        \State $\{(v_k,i_k)\}_{k\leq K}
        \leftarrow f_{i^\star}(\tau_j^-,\tau^+)$
        \State $\{m_k\}_{k\le K}=\{(\text{enforce } i_k;\ \text{avoid } v_k)\}_{k\le K}$
    \EndFor
    \State $\mathcal{M}^{(x)}\leftarrow\bigcup_{\tau_{l^-} \in \mathcal{T}^-} \left\{\left(\{m_k\}_{k=1}^K, \ell^+, \ell^-\right)\right\}$
    \State $\mathcal{M}\leftarrow \mathcal{M}\cup \mathcal{M}^{(x)}$
\EndFor

\State \textbf{return} $\mathcal{M}$
\end{algorithmic}

\hrule
\end{minipage}

\vspace{-2.5em}
\end{wrapfigure}
By abstracting from instance-specific solutions to reusable reasoning constraints, the extracted guidance can generalize across problems. These labels preserve the contrast direction, indicating which model provides the successful reference and which exhibits the failure pattern. This pairwise labeling supports model-aware retrieval and reduces interference among heterogeneous agents. The final memory bank $\mathcal{M}$ stores only \textbf{normative, abstract reasoning constraints} with preference-pair model labels, rather than raw trajectories, demonstrations, or final solutions.

\subsection{Task-Aware Memory Retrieval and Inference}
Given a new task $q$ and a target agent $A_j$, \ourmethod retrieves memory entries from the shared memory bank $\mathcal{M}$ and conditions the agent's inference on the retrieved constraints; see Algorithm~\ref{alg:retrieve} in the appendix. The retrieval follows a task-first and model-aware strategy. Specifically, we first retrieve memories that match the task type of $q$, and then further retain memories constructed from contrastive pairs in which the target agent $A_j$ participated, regardless of whether it produced the preferred or unpreferred trajectory.

\textbf{Task Categorization.}
Each memory entry in $\mathcal{M}$ is associated with the task category of its source task and the identities of the two agents involved in the corresponding contrastive pair. We denote a memory entry as
\begin{equation}
    m = (\eta_m, c_m, u_m, \ell_m^+, \ell_m^-),
\end{equation}
where $\eta_m$ denotes the stored memory content, i.e., the normative reasoning constraint of the form $(\text{enforce } i_k;\ \text{avoid } v_k)$, $(c_m,u_m)$ is the task category and subcategory inherited from the source task, and $(\ell_m^+,\ell_m^-)$ are the identities of the agents that produced the preferred and unpreferred trajectories, respectively. Given the test task $q$, a base LLM classifier assigns it a category and subcategory:
\begin{equation}
    (c_q,u_q)=f(q).
\end{equation}
\textbf{Task-Aware Filtering.}
We first restrict retrieval to memory entries whose task labels match the test task:
\begin{equation}
    \mathcal{M}_{q}^{\mathrm{task}}
    =
    \{m \in \mathcal{M}
    \mid
    c_m=c_q,\ u_m=u_q
    \}.
\end{equation}
This step ensures that the retrieved constraints are drawn from problems with similar reasoning structure, rather than from unrelated task domains.

\textbf{Model-Aware Filtering.}
After task-aware filtering, we further select memory entries whose contrastive construction involved the target agent. Let $\ell_j$ denote the identity label of $A_j$. A memory entry is retained if $\ell_j$ appears as either the preferred-model label or the unpreferred-model label:
\begin{equation}
    \mathcal{M}_{q,j}^{\mathrm{cand}}
    =
    \{m \in \mathcal{M}_{q}^{\mathrm{task}}
    \mid
    \ell_j \in \{\ell_m^+,\ell_m^-\}
    \}.
\end{equation}
This design allows the target agent to benefit from both directions of contrast. If $\ell_j=\ell_m^-$, the retrieved memory captures failure patterns that the target agent should avoid. If $\ell_j=\ell_m^+$, the retrieved memory captures reasoning invariants that the target agent has successfully followed and can reinforce. Thus, any memory constructed from a comparison involving $A_j$ is considered model-relevant.

\textbf{Relevance Ranking and Inference.}
Finally, \ourmethod ranks the candidate memories by their semantic relevance to the test task $q$ and selects the top-$p$ entries:
\begin{equation}
    \mathcal{M}_{q,j}
    =
    \mathrm{Top}\text{-}p
    \left(
    \mathcal{M}_{q,j}^{\mathrm{cand}},
    \mathrm{sim}(q,\eta_m)
    \right).
\end{equation}
The retrieved memory set $\mathcal{M}_{q,j}$ is then incorporated into the inference prompt of the target agent $A_j$. During generation, the agent is encouraged to follow the retrieved reasoning invariants and avoid the corresponding violation patterns. The detailed retrieval procedure is provided in Appendix~\ref{appendix:retrieval_algorithm}. This two-stage retrieval process first enforces task relevance and then model relevance. Thus the target agent receives constraints that are both appropriate for the current task and grounded in contrastive experience involving that agent, reducing interference from unrelated tasks or model comparisons.

\subsection{Why Contrast Works}
Constructing transferable memory across agents is challenging because raw reasoning trajectories entangle \textbf{task-relevant structure} with \textbf{agent-specific bias}, such as stylistic preferences and heuristic shortcuts. Memory distilled from a single agent may preserve such bias, limiting transferability. Suppose an agent trajectory is modeled as $\tau = f(s, b)$, where $s$ denotes the task-relevant reasoning structure and $b$ captures agent-specific bias. Since successful trajectories for the same task share similar $s$ despite varying $b$, collaborative memory aims to capture the invariant reasoning structure while suppressing agent-specific information.  Accordingly, we interpret the distilled memory as $m=\phi(s)$. By contrasting a preferred trajectory against an unpreferred one for the same task, \ourmethod highlights reasoning factors that distinguish success from failure while suppressing variations irrelevant to correctness. This intuition aligns with \textbf{contrastive learning}; Appendix~\ref{app:contra} provides a MemNCE-style interpretation of this mechanism.

\section{Experiments}
\label{sec:exp}

This section evaluates the effectiveness of our framework, \ourmethod, for constructing a single collaborative memory system that benefits multiple agents through contrastive memory learning.

\sisetup{
  table-number-alignment = center,
  table-format = 2.1,
  detect-weight = true,
  detect-inline-weight = math
}

\definecolor{Ours7B}{HTML}{E3E0F2}   
\definecolor{Ours32B}{HTML}{FFD6CF}  
\definecolor{HeaderGray}{HTML}{F4F4F4}

\begin{table*}[!t]
\centering
\small
\setlength{\tabcolsep}{5.5pt}
\renewcommand{\arraystretch}{1.15}

\begin{tabular}{ll
                S[table-format=2.1]
                S[table-format=2.1]
                S[table-format=2.1]
                S[table-format=2.1]
                S[table-format=2.1]}
\toprule
\rowcolor{HeaderGray}
\textbf{Backbone} &
\textbf{Method} &
\multicolumn{5}{c}{\textbf{Accuracy (\%) $\uparrow$}} \\
\cmidrule(lr){3-7}
& &
\textbf{MATH500} &
\textbf{GSM8K} &
\textbf{MBPP} &
\textbf{HumanEval} & 
\textbf{Average} \\
\midrule

\multirow{7}{*}{\textbf{Qwen-2.5-7B}}
& Vanilla
& 52.2 & 85.4 & 47.9 & 42.7 & 57.1 \\

& Buffer of Thoughts
& 45.8 & 86.4 & \textbf{57.6} & 62.2 & 63.0 \\

& Dynamic Cheatsheet
& 66.0 & 83.8 & 36.7 & 40.7 & 56.8 \\

& w/ Memory (7B)
& 60.0 & 86.2 & 50.2 & 42.7 & 59.8 \\

& w/ Memory (32B)
& 50.6 & 86.6 & 48.6 & 34.1 & 55.0 \\

& w/ Self-Contrast Memory
& 58.4 & 85.6 & 52.5 & 73.3 & 67.5 \\

\rowcolor{Ours7B}
& \textbf{\ourmethod}
& \textbf{67.0} & \textbf{87.4} & \textbf{57.6} & \textbf{74.4} & \textbf{71.6} \\
\midrule

\multirow{6}{*}{\textbf{Qwen-2.5-32B}}
& Vanilla
& 63.8 & 93.0 & 58.0 & 68.3 & 70.8 \\

& Buffer of Thoughts
& 68.2 & 93.0 & 63.8 & 84.1 & 77.3 \\

& Dynamic Cheatsheet
& 73.4 & 90.8 & 57.6 & 64.6 & 71.6 \\

& w/ Memory (32B)
& 69.6 & 93.2 & 60.3 & 79.3 & 75.6 \\

& w/ Self-Contrast Memory
& 69.6 & 93.4 & 58.7 & \textbf{87.8} & 77.4 \\

\rowcolor{Ours32B}
& \textbf{\ourmethod}
& \textbf{73.8} & \textbf{93.6} & \textbf{64.3} & 86.6 & \textbf{79.6} \\
\bottomrule
\end{tabular}
\vspace{1em}
\caption{
Performance of different memory methods with two models on math and code benchmarks.}
\vspace{-0.3em}

\label{tab:7b_results}
\end{table*}

\textbf{Models and Datasets.}
We conduct experiments using Qwen2.5-7B-Instruct~\citep{team2024qwen2}, Qwen2.5-32B-Instruct~\citep{team2024qwen2}, and LLaMA-3-8B-Instruct~\citep{grattafiori2024llama} as backbone models. Our evaluation covers two categories of reasoning tasks: mathematical reasoning and code generation. For mathematical reasoning, we use the MATH500~\citep{lightman2023let, math500} and GSM8K~\citep{cobbe2021training} datasets. From each dataset, we randomly sample 1000 instances to construct the memory system and evaluate performance on a disjoint set of 500 randomly selected instances, reporting accuracy as the metric~\cite{kang2025distilling}. For code generation, we evaluate on MBPP~\citep{austin2021program} and HumanEval~\citep{chen2021evaluating}. We explain more about the dataset experimental settings in Appendix~\ref{appendix:exp}.

\textbf{Baselines.}
We compare \ourmethod against two groups of baselines. 
\textbf{(i) Memory content baselines.} Buffer of Thoughts (BoT)~\citep{yang2024buffer} maintains a meta-buffer of reusable, high-level \emph{thought templates} distilled from prior problem-solving; at test time, it retrieves and instantiates a relevant template to guide reasoning, with a buffer manager for updates. Dynamic Cheatsheet (DC)~\citep{suzgun2025dynamic} equips an LLM with a persistent, evolving \emph{cheatsheet} at inference time, iteratively updated to store reusable guidance.
\textbf{(ii) Memory source baselines.} To isolate collaborative construction from naive transfer, we consider variants that differ only in where memory is derived: w/ Memory(7B) and w/ Memory(8B) use memory summarized from 7B or 8B agent trajectories, respectively; w/ Memory(32B) uses memory from 32B agent trajectories; and w/ Self-Contrast Memory constructs memory by contrasting multiple sampled trajectories from the same agent. Memory extraction follows BoT~\citep{yang2024buffer}. Detailed prompts are in Appendix~\ref{appendix:prompt}. 

\subsection{Main Results}
Table~\ref{tab:7b_results} reports results for agents instantiated with within the same backbone model family (Qwen-2.5) across different tasks, with variance reported in Appendix~\ref{appendix:variance}. Additional results across model families, scale regimes, and task settings are provided in Appendix~\ref{appendix:model_generalization}. The results on agentic planning task are provided in Appendix~\ref{sec:planning}

\textbf{Cross-model memory helps smaller-model-based agents benefit from larger-model trajectories.}
When equipped with \ourmethod, the agent using the smaller backbone model achieves substantial gains across all benchmarks, improving from 52.2\% to 67.0\% on MATH500 and from 47.9\% to 57.6\% on MBPP, approaching or even surpassing stronger baselines. These improvements indicate that contrast-derived memory captures transferable reasoning guidance across backbone models, rather than preserving model-specific behaviors.

\textbf{Contrastive memory benefits both weaker- and stronger-model-based agents.}
For the Qwen2.5-32B-based agent, \ourmethod consistently improves performance over the vanilla setting, achieving the highest accuracy on MATH500 (73.8\%), GSM8K (93.6\%), and MBPP (64.3\%). This demonstrates that collaborative memory does not merely compensate for weaker agents, but also enhances agents built on stronger backbone models by reducing systematic reasoning and tool-usage errors.

\textbf{Direct cross-model memory transfer can degrade performance.}
Using memory distilled solely from trajectories generated by the 32B-based agent (\textit{w/ Memory (32B)}) degrades the performance of the 7B-based agent on MATH500 (50.6\% vs.\ 52.2\%) and HumanEval (34.1\% vs.\ 42.7\%), highlighting the limitations of naively reusing memory across agents instantiated with different backbone models. In contrast, self-contrast memory yields partial improvements, but remains inferior to \ourmethod, underscoring the importance of cross-model contrast for filtering out model-specific biases while preserving reusable reasoning invariants.

\subsection{Results across Model Families}
Table~\ref{tab:8b_results} presents the results of \ourmethod when collaborative memory is constructed from agents using different backbone model architectures.

\sisetup{
  table-number-alignment = center,
  table-format = 2.1,
  detect-weight = true,
  detect-inline-weight = math
}

\definecolor{Ours7B}{HTML}{E3E0F2}   
\definecolor{Ours32B}{HTML}{FFD6CF}  
\definecolor{HeaderGray}{HTML}{F4F4F4}

\begin{table*}[!t]
\centering
\small
\setlength{\tabcolsep}{5.5pt}
\renewcommand{\arraystretch}{1.15}

\begin{tabular}{ll
                S[table-format=2.1]
                S[table-format=2.1]
                S[table-format=2.1]
                S[table-format=2.1]
                S[table-format=2.1]}
\toprule
\rowcolor{HeaderGray}
\textbf{Backbone} &
\textbf{Method} &
\multicolumn{5}{c}{\textbf{Accuracy (\%) $\uparrow$}} \\
\cmidrule(lr){3-7}
& &
\textbf{MATH500} &
\textbf{GSM8K} &
\textbf{MBPP} &
\textbf{HumanEval} &
\textbf{Average} \\
\midrule

\multirow{7}{*}{\textbf{Llama-3-8B}}
& Vanilla
& 27.4 & 73.0 & 37.0 & 29.3 & 41.7 \\

& Buffer of Thoughts
& 20.6 & 73.6 & 43.2 & 34.2 & 42.9 \\

& Dynamic Cheatsheet
& 29.4 & 71.2 & 27.2 & 17.1 & 36.2 \\

& w/ Memory (8B)
& 29.2 & 63.2 & 47.1 & 42.7 & 45.6 \\

& w/ Memory (32B)
& 18.8 & 56.6 & 35.8 & 34.1 & 36.3 \\

& w/ Self-Contrast Memory
& 23.6 & 61.0 & \textbf{51.4} & {43.9} & 45.0 \\

\rowcolor{Ours7B}
& \textbf{\ourmethod}
& \textbf{42.4} & \textbf{74.4} & 49.8 & \textbf{48.8} & \textbf{53.9} \\
\midrule

\multirow{3}{*}{\textbf{Qwen-2.5-32B}}
& Vanilla
& 63.8 & 93.0 & 58.0 & 68.3 & 70.8 \\

& w/ Memory (32B)
& 69.6 & 93.2 & \textbf{60.3} & 79.3 & 75.6 \\

\rowcolor{Ours32B}
& \textbf{\ourmethod}
& \textbf{70.6} & \textbf{95.2} & \textbf{60.3} & \textbf{86.6} & \textbf{78.2}\\
\bottomrule
\end{tabular}
\caption{
Performance of Llama3-8B-Instruct– and Qwen2.5-32B-Instruct–based agents on mathematical reasoning and code benchmarks.}
\label{tab:8b_results}
\end{table*}

\textbf{Contrastive memory generalizes across model architectures.}
With cross-family agent pairs, \ourmethod consistently improves performance across benchmarks. For instance, the Llama3-8B-based agent equipped with cross-family contrastive memory reaches 74.4\% on MATH500 and 48.8\% on HumanEval, outperforming both the vanilla baseline and single-source memory transfer. The Qwen2.5-32B-based agent also benefits despite architectural differences. These results suggest that contrastive memory captures reusable reasoning strategies across agents with different backbone model families, rather than model-family-specific behaviors.

\subsection{Robustness to Heterogeneous Memory Sources.}

\begin{wraptable}{r}{0.5\textwidth}
\vspace{-1em}
\centering
\vspace{-0.5em}
\resizebox{\linewidth}{!}{
\begin{tabular}{c|ccc}
\toprule
\# Models & Qwen2.5-7B & Llama3-8B & Qwen2.5-32B \\
\midrule
1 & 52.2 & 27.4 & 58.0 \\
2 & \textbf{67.0} & 42.4 & 73.8 \\
3 & 66.8 & \textbf{43.2} & \textbf{75.0} \\
\bottomrule
\end{tabular}
}
\caption{Three-model collaboration on MATH500.}
\label{tab:three_agent_math500}
\end{wraptable}

Table~\ref{tab:three_agent_math500} examines whether incorporating memory distilled from an additional heterogeneous backbone introduces interference into the shared bank. We compare the original two-agent memory bank constructed from Qwen2.5-7B and Qwen2.5-32B with a variant that also includes trajectories from Llama3-8B, and evaluate the same memory-augmented inference procedure on MATH500.
The results show that the shared bank remains stable under this added source of heterogeneity. The newly included Llama3-8B agent also benefits from the shared bank, improving from 42.4 to 43.2. These results indicate that additional heterogeneous trajectories do not automatically introduce harmful cross-agent interference.

The non-uniform gains are also informative. The added Llama3-8B agent lies closer in capability to Qwen2.5-7B than to Qwen2.5-32B, and therefore may contribute failure patterns that partially overlap with those already captured for the 7B agent. This explains why Qwen2.5-7B does not obtain an additional gain, while Qwen2.5-32B benefits from the newly introduced contrast. Overall, Table~\ref{tab:three_agent_math500} supports the robustness of \ourmethod’s shared-bank, model-aware-retrieval design under heterogeneous memory sources, rather than suggesting that performance should increase simply by adding more agents.

\subsection{Effect of Memory Retrieval}
\label{sec:memory_retrieval}

We study the effect of memory retrieval from two perspectives: the number of retrieved memory entries and the strategy used to select relevant memories. Additional ablation studies, including the effects of summarizer choice, prompt stability, and preference selection, are provided in Appendix~\ref{appendix:more_ablation}.

\begin{figure}[t]
    \centering
    \begin{minipage}[t]{0.48\textwidth}
        \centering
        \includegraphics[width=\linewidth]{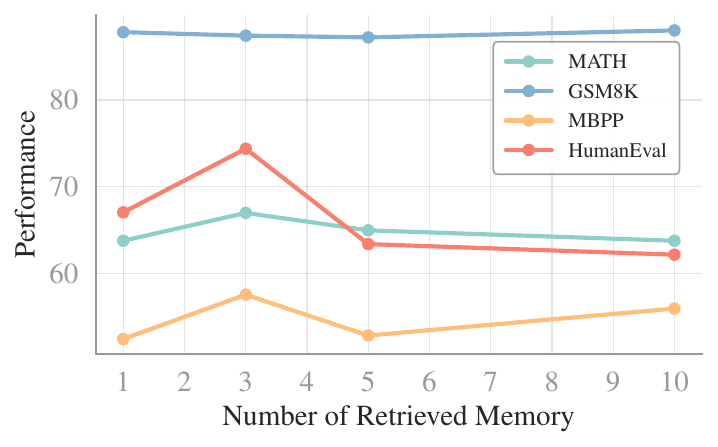}
        \caption{Ablation on the number of retrieved memories.}
        \label{fig:num_of_mem}
    \end{minipage}
    \hfill
    \begin{minipage}[t]{0.48\textwidth}
        \centering
        \includegraphics[width=\linewidth]{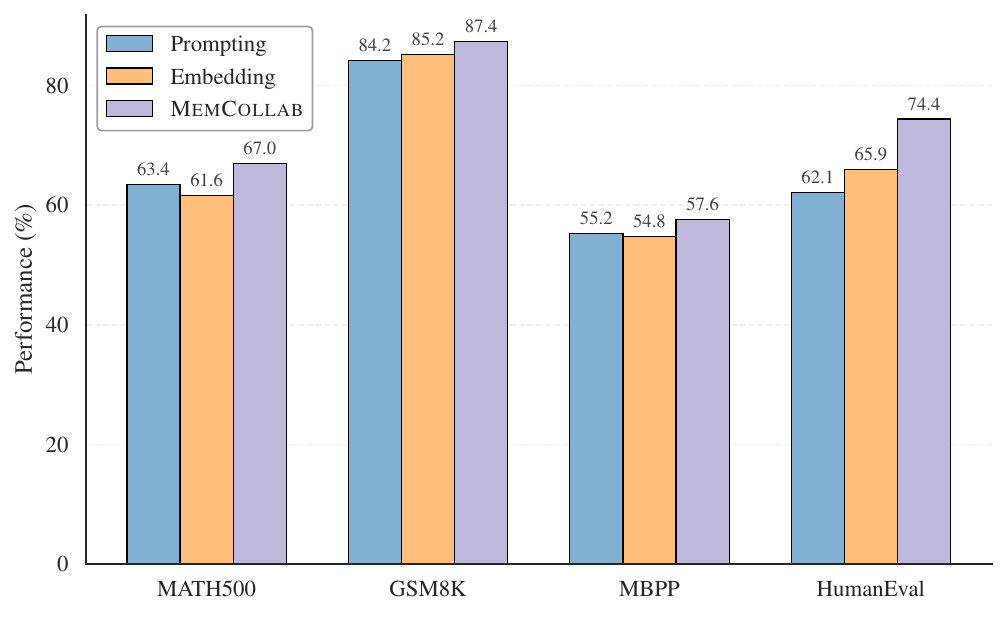}
        \caption{Performance comparison with different retrieval methods.}
        \label{fig:retrieve_method}
    \end{minipage}
    \vspace{-1em}
\end{figure}

\textbf{Number of retrieved memory entries.}
Figure~\ref{fig:num_of_mem} shows that performance first improves as more memories are retrieved, but drops beyond a task-dependent threshold. Since retrieving the top three memories consistently yields strong results, we set $p=3$ in all experiments. This trend can be viewed as a pruning trade-off. Retrieved contrastive memories help eliminate irrelevant reasoning directions and improve the signal-to-noise ratio. If $b$ is the original branching factor, $d$ is the search depth, and $k$ erroneous patterns are avoided, the effective search space contracts by
$
\rho = \left(1 - \frac{k}{b}\right)^d .
$
When $k \ge 1$, $\rho < 1$, indicating reduced search complexity. However, retrieving too many memories introduces weakly relevant or noisy directions, causing attention dispersion and weakening the pruning effect. This explains the non-monotonic trend and motivates a task-dependent retrieval budget.

\textbf{Retrieval strategy.}
We also evaluate two retrieval variants that do not use task classification, instead retrieving memory entries directly based on (1) embedding similarity and (2) prompting the model to select the top-$p$ relevant tasks. As shown in Figure~\ref{fig:retrieve_method}, these results highlight the benefit of first classifying the task and then retrieving memory conditioned on the predicted class. Compared with retrieval over the full memory bank, task-aware retrieval narrows the candidate space and reduces interference from memories associated with unrelated failure patterns.

\subsection{Error-Pattern Analysis by Task Category}
\label{sec:error_pattern_analysis}

\begin{wrapfigure}{r}{0.46\textwidth}
    \vspace{-1em}
    \centering
    \includegraphics[width=0.44\textwidth]{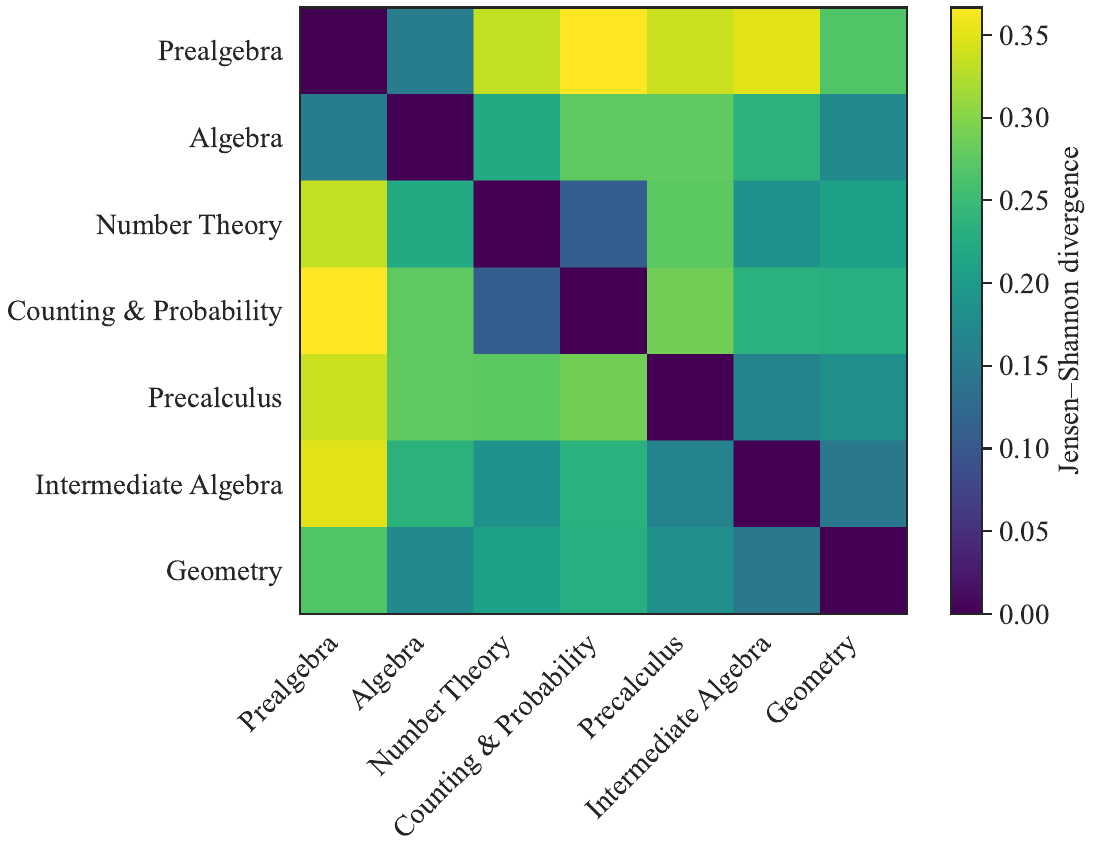}
    \caption{Task-category similarity of error-type distributions on MATH500.}
    \label{fig:error-heatmap}
\end{wrapfigure}

To demonstrate the relationship between task categories and error patterns, we analyze the pairwise Jensen--Shannon divergence (JSD) between task categories on the MATH500 dataset using the Qwen2.5-7B model, as shown in Figure~\ref{fig:error-heatmap}. Each task category is represented by an empirical distribution over error types aggregated from model-generated trajectories. For each data sample and its corresponding trajectory, we use GPT-4o-mini to annotate both the task category and the error type, and compute the JSD between all pairs of categories.

\textbf{Task categories exhibit characteristic and consistent error patterns.}
The results in Figure~\ref{fig:error-heatmap} reveal that algebraically related tasks, including Prealgebra, Algebra, Intermediate Algebra, and Precalculus, form a low-divergence cluster, indicating highly similar error distributions dominated by reasoning and execution-related failures. In contrast, categories such as Number Theory and Counting \& Probability deviate from this cluster, indicating distinct failure mechanisms. These findings support the design of our task-aware retrieval mechanism, as memory entries derived from the same task category are more likely to preserve relevant reasoning constraints and error-forbidden patterns, enabling more targeted guidance during inference.

\subsection{Case Study}
\begin{figure*}[!h]
    \centering
    \includegraphics[width=\textwidth]{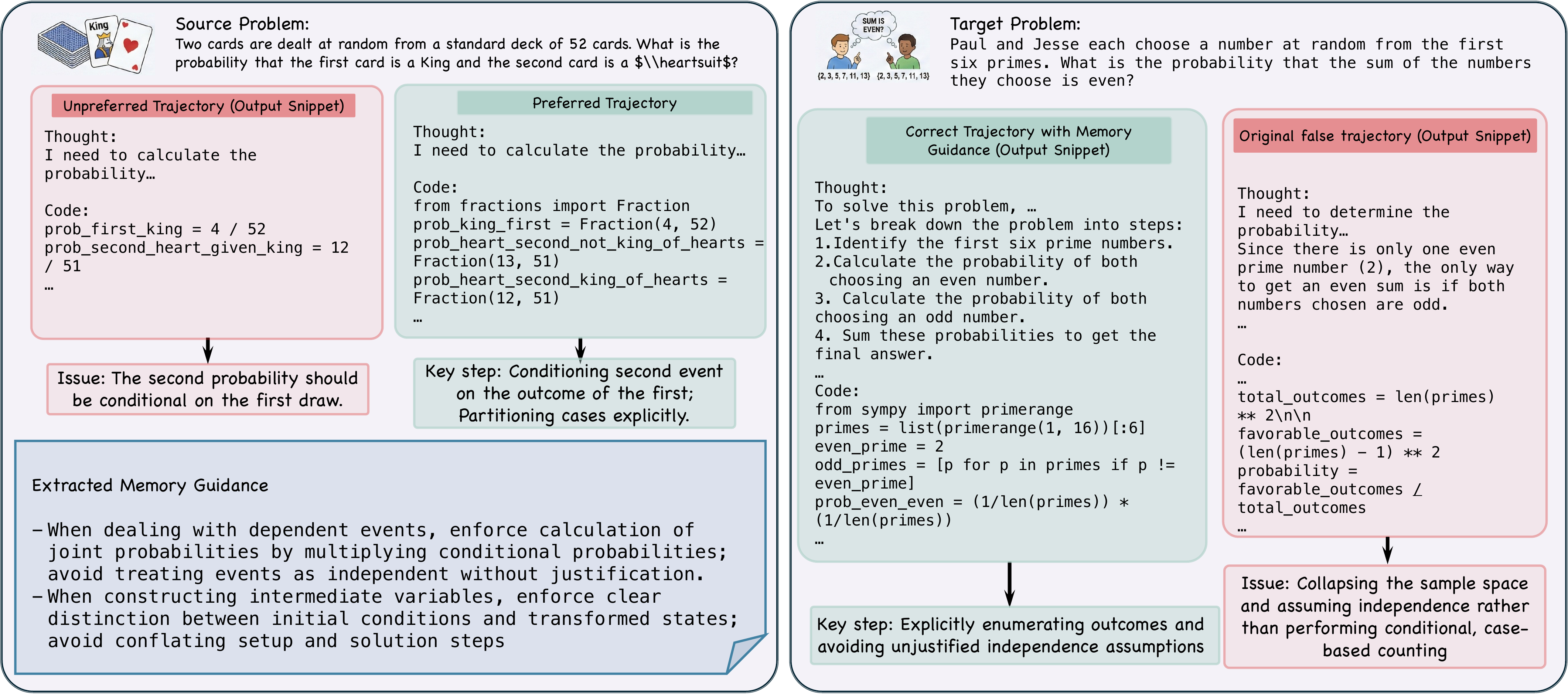}
    \caption{Case study to compare the trajectories with and without the extracted memory guidance. \textit{Left}: The contrasted trajectories pairs. \textit{Right}: Changes in the agent's output after incorporating the memory guidance. Note: all \texttt{Key step} and \texttt{Issue} are summarized, see full output is in Appendix~\ref{appendix:full_traj}.}
    \label{fig:case_study}
\end{figure*}

In Figure~\ref{fig:case_study}, we present a case on how \ourmethod works. For the source problem, the unpreferred response mistakenly assumes independence in a setting with dependencies in a probability problem. In contrast, the preferred response correctly models the joint probability by conditioning on earlier events and systematically enumerating the relevant cases. The resulting memory guidance distills this central distinction between the two reasoning trajectories. When solving the target problem, the vanilla agent exhibits the same failure mode, again relying on an implicit independence assumption and an underspecified enumeration of cases. After retrieving the contrastive memory, the agent can adopt explicit case analysis, leading to a coherent joint-probability formulation and the correct solution. Appendix~\ref{appendix:memory_quality_judge} also shows that \ourmethod{} produces higher-quality memory than Self-Memory under an LLM-as-a-judge evaluation.

\section{Related Works}

\textbf{Memory for LLM Agents}
Memory has become a core module for LLM-based agents, helping them mitigate limited context windows and reuse information across episodes~\citep{zhang2025survey, nuxoll2012enhancing}. Stored memory may appear as free-form text or structured artifacts and can be retrieved to guide later reasoning and tool use~\citep{ouyang2025reasoningbank, zhong2024memorybank, packer2023memgpt}. Existing mechanisms include episodic memory for recording interactions and managing multi-tier context~\citep{packer2023memgpt}, long-term interaction memory for continual recall and update~\citep{zhong2024memorybank}, and reasoning-oriented memory that distills reusable problem-solving guidance~\citep{yang2024buffer, suzgun2025dynamic, ouyang2025reasoningbank}. Other studies focus on memory organization and operations, such as evolving memory graphs~\citep{xu2025mem} and training-based policies for storage, update, and retrieval~\citep{zhou2025mem1, yan2025memory, yang2025reasonflux}. Despite their effectiveness, these approaches usually target a single agent: memory is synthesized from one agent's trajectories and reused by the same agent. This limits cross-agent transfer. In contrast, our work studies how to construct a shared memory system for agents with different model sizes and architectures. Rather than storing episodic traces or learning agent-specific memory policies, we contrast trajectories from different agents on the same task to distill transferable reasoning constraints and error-forbidden patterns.

\textbf{Retrieval-Augmented Model Generation}
Retrieval-augmented generation (RAG) extends LLMs with external corpora and improves factuality and coverage by conditioning outputs on retrieved passages~\citep{lewis2020retrieval}. Prior work shows that retrieval source and unit granularity strongly affect generation quality~\citep{gao2023retrieval}. Retrieval sources have expanded from unstructured text~\citep{li2023classification, yan2024corrective} to semi-structured~\citep{zha2023tablegpt, luo2023augmented} and structured data~\citep{gaur2022iseeq, he2024g, edge2024local}, as well as LLM-generated or distilled content~\citep{yu2022generate, cheng2023lift}. Retrieval granularity also ranges from full documents or long passages~\citep{yu2024chain, shi2023large} to finer units such as sentences, entities, or code snippets~\citep{chen2024dense, wang2024learning}. Unlike document-centric RAG, our method retrieves reasoning-oriented memory rather than external knowledge. We use coarse-grained memory units but apply task-level classification before retrieval, restricting the search space to relevant memory and reducing incompatible guidance. This enables efficient retrieval of task-relevant reasoning constraints without the noise of overly fine-grained retrieval.

\section{Conclusion}
In this work, we study shared cross-model memory for LLM-based agents. We show that naively transferring memory across agents with different backbone models can degrade performance, as memory may entangle reusable knowledge with model-specific biases. To address this, we propose \ourmethod{}, which constructs collaborative memory by contrasting trajectories generated by different model-based agents on the same task. Experiments on mathematical reasoning and code generation benchmarks show that \ourmethod{} improves both accuracy and inference-time efficiency across heterogeneous agents. These results suggest that contrast-derived memory can serve as a reusable reasoning resource for cross-model agent systems.

\bibliographystyle{plainnat}
\bibliography{contents/reference}

\clearpage
\appendix

\section{Additional Theoretical Details}
\label{sec:appendix_theory}
This appendix provides supplementary theoretical analysis to support our interpretation of our contrastive memory construction. 

\subsection{Contrastive Learning}
\label{app:contra}

Contrastive learning provides a statistical framework for understanding how representations can be shaped by the comparison between the positive and negative pairs. The widely used contrastive objective loss function is the infoNCE loss:
\begin{equation}
\mathcal{L}_{\mathrm{InfoNCE}}
=
-\mathbb{E}\!\left[
\log
\frac{\exp(g(x,x^+))}
{\exp(g(x,x^+)) + \sum_j \exp(g(x,x_j^-))}
\right],
\end{equation}

where $g(\cdot, \cdot)$ denotes the scoring function, $x$ and $x^+$ are from positive pairs, and $\{x_j^-\}$ are negative samples~\citep{logeswaran2018efficient}. 

The simplified version of the loss function could be defined as~\cite{saunshi2019theoretical}:

\begin{equation}
-\mathbb{E}\!\left[
\log
\frac{\exp(g(x,x^+))}
{\exp(g(x,x^+)) + \exp(g(x,x^-))}
\right],
\end{equation}

\subsection{Contrastive Memory Learning}
We analyze the behavior of the contrast-derived memory from the perspective of the contrastive learning. 

Suppose an agent trajectory is represented as $\tau$, which is composed of two factors: a task-specific reasoning strategies and agent-specific bias. Formally, we assume:
\begin{equation}
\tau = f(s, b),
\end{equation}
where $s$ denotes the underlying reasoning structure required to solve the task, and $b$ captures agent-dependent variations such as architectural preferences or stylistic choices. Accordingly, we interpret the contrast-derived memory $m$ as a function of the reasoning component, i.e., $m=\psi(s)$, while being largely insensitive to the bias term $b$. Given these assumputions, we define a proxy optimization objective for the contrastive memory learning as:
\begin{equation}
\mathcal{L}_{\mathrm{MemNCE}}
=
-\mathbb{E}\!\left[
\log
\frac{\exp(g(\tau,m))}
{\exp(g(\tau,m)) + \sum_j \exp(g(\tau_j^-,m))}
\right],
\end{equation}
where $\{\tau_j^-\}$ denote false trajectories corresponding to the same task. In our setting, contrast is performed over a single preferred trajectory and a single flawed trajectory, corresponding to the pairwise case with $j=1$ negative sample. \citeauthor{oord2018representation} shows that even with a single negative sample,
pairwise contrast preserves the density-ratio interpretation central to contrastive learning.

Under this view, the $\mathcal{L}_{\mathrm{MemNCE}}$ provides an interpretive tool for understanding the effectiveness of the contrast-derived memory. Instead of directly optimizing this objective, our goal is to introduce a proxy learning objective that clarifies the underlying memory extraction mechanism. By contrasting preferred and unlike trajectories, the distilled memory emphasizes reasoning factors and recurring pitfalls that persist across problem-solving processes, while ignoring the agent-specific biases and noises. Unlike episodic storage or direct trajectory reuse, our collaborative memory encourages memory distillation into abstract reusable, high-level guidance. This perspective helps explain why memory constructed through contrast can generalize across tasks and agents. 

\section{Additional Experimental Details}
\label{appendix:exp}
All experiments are conducted on 8 NVIDIA RTX 6000 GPUs with CUDA 12.8. We adopt smolagents’ CodeAgent as the underlying agent framework. During inference, we set the temperature to 0 and top-p to 0.8. For memory retrieval, we retrieve the top-3 most similar samples within the task-filtered subset using TF-IDF similarity scores; each retrieved sample contributes up to 3 extracted memory entries, which are then provided to the agent. For the self-contrast baseline, each model generates two sampled outputs to construct contrastive memory. We use the random seed 42 for all experiments. 
\subsection{Dataset}
\label{appendix:data}
\begin{itemize}[leftmargin=*]
    \item \textbf{MATH500}: A curated reasoning benchmark consisting of 500 problems sampled from the MATH dataset~\citep{math500} following the protocol of \citet{lightman2023let}. The benchmark spans seven mathematical domains, including Prealgebra, Algebra, Number Theory, Counting and Probability, Geometry, Intermediate Algebra, and Precalculus, ensuring to provide diverse coverage of problem types and reasoning requirements. 
    
    \item \textbf{GSM8K} A grade-school mathematics benchmark consisting of 8,500 natural-language word problems that require multi-step arithmetic reasoning~\citep{cobbe2021training}. Each problem is paired with a step-by-step solution and a final numeric answer, and the dataset is commonly used to evaluate an agent’s ability to perform structured mathematical reasoning. 

    \item \textbf{MBPP} (Mostly Basic Programming Problems): A benchmark for evaluating code generation and program synthesis, consisting of 974 Python programming tasks~\citep{austin2021program}. Each task provides a natural-language problem description, a reference solution, and a set of test cases, and is designed to assess an agent’s ability to generate correct and executable code from textual specifications.

   \item \textbf{HumanEval}: A widely used benchmark for evaluating code generation and program synthesis capabilities of large language models, consisting of 164 hand-written Python programming tasks~\citep{chen2021evaluating}. Each task specifies a function signature and a natural-language docstring, and models are evaluated based on functional correctness using unit tests, typically reported with pass@k metrics.

\end{itemize}
\subsection{Baselines}
\label{appendix:baselines}
\begin{itemize}[leftmargin=*]
    \item Buffer of Thoughts~\citep{yang2024buffer}: proposes a reasoning-oriented memory mechanism that distills reusable thought templates from prior problem-solving traces. During inference, the agent retrieves and instantiates these templates to guide its reasoning process, aiming to improve consistency and efficiency on complex reasoning tasks.
    \item Dynamic Cheatsheet~\citep{suzgun2025dynamic}: introduces an inference-time memory mechanism that maintains a persistent, editable cheatsheet of task-relevant guidance. The cheatsheet is updated online based on model successes and failures, enabling black-box language models to adapt their behavior across tasks without parameter updates.
    \item Vanilla: is to use the basic agent wthout memory augmentation.
    \item w/ Memory (7B): is to utilize the memory constructed exclusively from reasoning trajectories generated by Qwen2.5-7B. During inference, the model retrieves the stored memory and conditions its generation on the retrieved content to guide problem solving. 
    \item w/ Memory (32B): has the setting mirrors w/ Memory (7B), except that the memory is constructed from reasoning trajectories generated by Qwen2.5-32B instead of Qwen2.5-7B. 
    \item w/ Self-Contrast Memory: constructs memory by contrasting the model’s own self-generated reasoning trajectories, using the resulting memory to guide inference.

\end{itemize}

\section{Additional Experiments}
\label{appendix:model_generalization}

We further examine whether \ourmethod{} generalizes across model families, scale regimes, and task settings. Rather than exhaustively covering all possible model combinations, this evaluation tests representative axes of heterogeneity beyond the main experiments, including larger-scale cross-family collaboration, additional cross-family collaboration, comparable-scale collaboration, larger comparable-scale collaboration, and long-horizon agentic planning.

These settings clarify that the gains of \ourmethod{} do not rely on a single backbone family, a single scale regime, or a large weak--strong capability gap. Since \ourmethod{} extracts reusable reasoning invariants and failure-avoidance constraints by contrasting reasoning trajectories, its effectiveness depends on useful cross-agent contrast rather than model size differences alone. Taken together, the results show that \ourmethod{} remains effective across larger-scale, comparable-scale, cross-family, and GPT-based agentic settings.

\subsection{Larger-scale cross-family collaboration.}
We first evaluate Llama-3-70B-Instruct in collaboration with Qwen2.5-7B, extending the study to a larger-scale Llama model. As shown in Table~\ref{tab:llama70b_qwen7b}, \ourmethod improves Llama-3-70B over both the vanilla and self-memory baselines across all benchmarks.

\begin{table}[h]
\centering
\begin{tabular}{llcccc}
\toprule
Model & Method & MATH500 & GSM8K & MBPP & HumanEval \\
\midrule
Llama-3-70B & Vanilla & 58.0 & 93.2 & 54.8 & 64.6 \\
             & w/ Memory (self) & 59.6 & 93.5 & 56.3 & 68.6 \\
             & w/ \ourmethod & \textbf{65.2} & \textbf{93.6} & \textbf{63.0} & \textbf{74.4} \\
\bottomrule
\end{tabular}
\vspace{4pt}
\caption{Larger-scale cross-family collaboration with Llama-3-70B-Instruct.}
\label{tab:llama70b_qwen7b}
\end{table}

\subsection{Additional cross-family collaboration.}
We also evaluate Gemma3-4B-Instruct with Qwen2.5-7B on MBPP, providing another cross-family setting beyond Qwen and Llama. As shown in Table~\ref{tab:gemma_qwen}, \ourmethod improves both Gemma3-4B-Instruct and Qwen2.5-7B-Instruct over their vanilla and self-memory baselines.

\begin{table}[h]
\centering
\begin{tabular}{lccc}
\toprule
Model & Vanilla & w/ Memory (self) & w/ \ourmethod \\
\midrule
Gemma3-4B & 45.9 & 45.1 & \textbf{48.4} \\
Qwen2.5-7B & 47.9 & 50.2 & \textbf{50.8} \\
\bottomrule
\end{tabular}
\vspace{4pt}
\caption{Additional cross-family collaboration between Gemma3-4B-Instruct and Qwen2.5-7B-Instruct on MBPP.}
\label{tab:gemma_qwen}
\end{table}

\subsection{Comparable-scale collaboration.}

\begin{table}[h]
\centering
\begin{tabular}{llcccc}
\toprule
Model & Setting & MATH500 & GSM8K & MBPP & HumanEval \\
\midrule
Qwen2.5-7B
& Vanilla & 52.2 & 85.4 & 47.9 & 42.7 \\
& w/ Memory (self) & 60.0 & \textbf{86.2} & 50.2 & 42.7 \\
& w/ \ourmethod & \textbf{63.8} & 85.4 & \textbf{55.3} & \textbf{71.2} \\
\midrule
Llama-3-8B
& Vanilla & 27.4 & 73.0 & 37.0 & 29.3 \\
& w/ Memory (self) & 29.2 & 63.2 & 47.1 & 42.7 \\
& w/ \ourmethod & \textbf{30.4} & \textbf{73.4} & \textbf{47.9} & \textbf{45.5} \\
\bottomrule
\end{tabular}
\vspace{4pt}
\caption{Comparable-scale cross-family collaboration between Qwen2.5-7B-Instruct and Llama-3-8B-Instruct.}
\label{tab:comparable_scale}
\end{table}

We further test whether \ourmethod remains useful when collaborating agents have similar overall capacity in Table~\ref{tab:comparable_scale}. This setting is important because it rules out the possibility that the gains come only from transferring knowledge from a much larger model to a much smaller one. In the comparable-scale cross-family setting, \ourmethod improves both Qwen2.5-7B-Instruct and Llama-3-8B-Instruct over their vanilla and self-memory baselines, showing that useful contrast still exists between similar-scale but architecturally different agents.

We additionally evaluate a larger comparable-scale setting with Llama-3-70B-Instruct and Qwen2.5-72B-Instruct on MBPP. As shown in Table~\ref{tab:large_comparable_scale}, \ourmethod improves Llama-3-70B from $54.8$ to $61.9$ and Qwen2.5-72B from
$59.9$ to $63.0$, outperforming both vanilla and self-memory baselines.
\begin{table}[t]
\centering
\begin{tabular}{llc}
\toprule
Model & Setting & Accuracy \\
\midrule
Llama-3-70B
& Vanilla & 54.8 \\
& w/ Memory (self) & 56.3 \\
& w/ \ourmethod & \textbf{61.9} \\
\midrule
Qwen2.5-72B
& Vanilla & 59.9 \\
& w/ Memory (self) & 61.1 \\
& w/ \ourmethod & \textbf{63.0} \\
\bottomrule
\end{tabular}
\vspace{4pt}
\caption{Large comparable-scale collaboration between Llama-3-70B and
Qwen2.5-72B on MBPP.}
\label{tab:large_comparable_scale}
\end{table}

\subsection{Generalization to Agentic Planning Tasks}
\label{sec:planning}
Beyond math and coding benchmarks, we additionally evaluate \ourmethod on AppWorld~\citep{trivedi2024appworld}, a challenging long-horizon multi-app agentic benchmark. This setting introduces additional challenges such as environment interaction, tool-use reliability, and
multi-step planning. Since the original 32B setup achieves near-zero performance on this benchmark, we instead evaluate GPT-5-mini and GPT-5-nano. As shown in Table~\ref{tab:appworld}, \ourmethod improves both task goal completion and
scenario goal completion for both GPT-based models.
\begin{table}[h!]
\centering
\resizebox{0.9\textwidth}{!}{
\begin{tabular}{ll|cccc|cccc}
\toprule
\multirow{2}{*}{Model} & \multirow{2}{*}{Method}
& \multicolumn{4}{c|}{Task Goal Completion}
& \multicolumn{4}{c}{Scenario Goal Completion} \\
\cmidrule(lr){3-6} \cmidrule(lr){7-10}
& & Avg. & D1 & D2 & D3 & Avg. & D1 & D2 & D3 \\
\midrule
GPT-5-nano & Vanilla
& 4.8 & 11.1 & 0.0 & 0.0
& 0.0 & 0.0 & 0.0 & 0.0 \\
& \ourmethod
& \textbf{26.7} & \textbf{44.4} & \textbf{26.7} & 0.0
& \textbf{10.0} & 0.0 & \textbf{20.0} & 0.0 \\
\midrule
GPT-5-mini & Vanilla
& 47.7 & 66.7 & 55.3 & 0.0
& 30.0 & 66.7 & 20.0 & 0.0 \\
& \ourmethod
& \textbf{63.3} & 66.7 & \textbf{66.7} & \textbf{50.0}
& \textbf{50.0} & 66.7 & \textbf{40.0} & \textbf{50.0} \\
\bottomrule
\end{tabular}%
}
\caption{Results on AppWorld. Task Goal Completion measures success on individual tasks, while Scenario Goal Completion is stricter and requires solving all variants within a scenario; D1–D3 denote increasing difficulty levels.}
\label{tab:appworld}
\end{table}

\subsection{Extension to the Open-ended Task}
Our method requires a correctness signal only during offline memory construction, where the indicator identifies preferred/unpreferred trajectories. At inference on new tasks, \ourmethod does not rely on ground truth; it only retrieves abstract reasoning constraints distilled from prior tasks as soft guidance. In this paper we focus on math/code because ground-truth answers or executable verifiers provide a clean testbed. The same framework can extend to more open-ended tasks by replacing the current indicator with other preference sources, e.g., human preferences, model judges, or task-specific proxy rewards. To test this, we additionally evaluate on ASQA, using the backbone model itself to select the preferred trajectory and the results are shown in the Table~\ref{tab:open_eneded}. This suggests the framework can extend beyond closed-form tasks when a task-level preference signal is unavailable.
\begin{table}[t]
\centering
\begin{tabular}{lcc}
\toprule
Model & Vanilla & w/ \ourmethod \\
\midrule
Qwen2.5-7B  & 15.4 & \textbf{21.3} \\
Qwen2.5-32B & 15.9 & \textbf{33.3} \\
\bottomrule
\end{tabular}
\vspace{2pt}
\caption{Performance comparison with and without \ourmethod on open-ended task ASQA.}
\label{tab:open_eneded}
\end{table}

\section{Additional Studies}

In this section, we provide additional analysis of \ourmethod.
\label{appendix:more_ablation}
\subsection{Effect of Preference Selection}
\label{appendix:preference_selection}

We further test whether the preference-selection rule affects the quality of
the constructed memory. On Qwen2.5-7B, we compare the default rule with two alternative strategies: \emph{random}, which randomly chooses the preferred trajectory, and \emph{reverse}, which swaps the preferred and unpreferred trajectories under the default rule.

As shown in Table~\ref{tab:preference_selection}, both alternatives still improve over the vanilla baseline, but they are clearly worse than the default selection rule. \ourmethod with the default rule achieves $57.6$ on MBPP,
compared with $50.2$ for random selection and $52.9$ for reverse selection. This indicates that correctly identifying the preferred trajectory is important for producing higher-quality memory.
\begin{table}[h!]
\centering
\begin{tabular}{lc}
\toprule
Method & MBPP \\
\midrule
Vanilla & 47.9 \\
MemCollab (default) & \textbf{57.6} \\
MemCollab (random) & 50.2 \\
MemCollab (reverse) & 52.9 \\
\bottomrule
\end{tabular}
\caption{Effect of preference selection on MBPP with Qwen2.5-7B.}
\label{tab:preference_selection}
\end{table}

\subsection{Effect of Summarizer Choice}
\label{appendix:summarizer_effect}

Algorithm~\ref{alg:construct} uses the stronger backbone for trajectory summarization. To test whether the gains depend on this specific summarizer, we vary only the summarizer model while keeping the rest of the MemCollab pipeline fixed. As shown in Table~\ref{tab:summarizer_effect}, summarizer capability affects memory quality: using Qwen2.5-32B-Instruct gives the best HumanEval result
($74.4$). However, the improvement is not specific to the larger model. Using
Qwen2.5-7B-Instruct or Llama-3-8B-Instruct as the summarizer still substantially improves over the
vanilla baseline, reaching $65.9$ and $63.4$, respectively. These results suggest
that stronger summarizers can produce higher-quality memory, but \ourmethod does
not rely exclusively on the strongest backbone.

\begin{table}[h]
\centering
\begin{tabular}{llc}
\toprule
Target Model & Summarizer & HumanEval \\
\midrule
Vanilla & -- & 42.7 \\
\midrule
\multirow{3}{*}{w/ \ourmethod}
& Qwen2.5-32B & \textbf{74.4} \\
& Qwen2.5-7B & 65.9 \\
& Llama-3-8B & 63.4 \\
\bottomrule
\end{tabular}
\vspace{4pt}
\caption{Effect of summarizer choice on HumanEval with Qwen2.5-7B as the target
model.}
\label{tab:summarizer_effect}
\end{table}

\subsection{Prompt Sensitivity}
\label{appendix:prompt_sensitivity}

We evaluate the sensitivity of \ourmethod to the summarization prompt. We keep
the same trajectory pairs and downstream pipeline, and vary only the prompt used
to summarize contrastive trajectories into memory entries. As shown in Table~\ref{tab:prompt_sensitivity}, the original prompt achieves $74.4$ on HumanEval, and adding an instruction to output a numbered list achieves the same performance. Enforcing a JSON output format slightly reduces performance to $70.7$, but still substantially improves over the no-memory baseline of $42.7$. These results suggest that the gain mainly comes from the distilled memory content rather than a brittle prompt template.

\begin{table}[h]
\centering
\begin{tabular}{lc}
\toprule
Prompt Change & HumanEval \\
\midrule
No memory & 42.7 \\
Original prompt & \textbf{74.4} \\
Add instruction to numbered list & \textbf{74.4} \\
Add instruction to JSON format & 70.7 \\
\bottomrule
\end{tabular}
\vspace{4pt}
\caption{Prompt sensitivity on HumanEval.}
\label{tab:prompt_sensitivity}
\end{table}

\subsection{Efficiency Analysis}
\label{appendix:memory_cost}

\paragraph{Memory Construction Cost} We analyze the memory construction cost on MBPP using Qwen3-32B as the summarizer. As shown in Table~\ref{tab:memory_cost}, \ourmethod requires more cost than self-memory, but is cheaper than self-contrast memory in both wall-clock time and total token usage. Specifically, \ourmethod uses $38.3$ seconds and $929.90$ total tokens on average, compared with $50.4$ seconds and $950.50$
tokens for self-contrast. Although self-memory is cheaper, \ourmethod provides a
stronger trade-off between construction cost and downstream performance in our
main experiments.

\begin{table}[h]
\centering
\begin{tabular}{lcccc}
\toprule
Method & Wall Time (s) & Input Tokens & Output Tokens & Total Tokens \\
\midrule
\ourmethod & 38.3 & 812.42 & 117.50 & 929.90 \\
Self-Contrast & 50.4 & 828.90 & 121.56 & 950.50 \\
Self-Memory & 13.35 & 499.02 & 63.78 & 562.80 \\
\bottomrule
\end{tabular}
\vspace{4pt}
\caption{Memory construction cost on MBPP using Qwen3-32B as the summarizer. }
\label{tab:memory_cost}
\end{table}

\begin{wraptable}{r}{0.5\textwidth}
\centering
\small
\setlength{\tabcolsep}{10pt}
\renewcommand{\arraystretch}{1.15}
\begin{tabular}{lcc}
\toprule
\textbf{Dataset} & \textbf{Vanilla} & \textbf{\ourmethod} \\
\midrule
MATH500     & 2.7 & 2.2 \\
GSM8K       & 1.8 & 1.6 \\
MBPP        & 3.1 & 1.4 \\
HumanEval   & 3.3 & 1.5 \\
\bottomrule
\end{tabular}
\caption{Average number of reasoning turns for Qwen2.5-7B-Instruct based agent.}
\label{tab:turns_7b}
\end{wraptable}

\paragraph{Memory improves inference-time efficiency.} We evaluate inference-time efficiency by comparing the average number of reasoning turns required by a vanilla agent and by \ourmethod. As shown in Table~\ref{tab:turns_7b}, agents equipped with collaborative memory consistently require fewer reasoning turns while achieving better overall performance. It indicates that contrast-derived memory provides effective guidance that reduces redundant exploration and unnecessary trial-and-error during inference. By discouraging previously observed failure patterns and instructing valid reasoning procedures, memory enables agents to converge to correct solutions more efficiently. 

\subsection{Variance Across Random Seeds}
\label{appendix:variance}

We report the standard deviation over three random seeds for both the vanilla
baseline and \ourmethod. As shown in Table~\ref{tab:variance}, \ourmethod
improves average performance across all benchmarks while maintaining comparable
or lower variance on most tasks. These results suggest that collaborative memory not only improves accuracy, but also helps stabilize trajectory-sensitive reasoning and implementation.
\begin{table}[h!]
\centering
\begin{tabular}{lcccc}
\toprule
Method & MATH500 & GSM8K & MBPP & HumanEval \\
\midrule
Vanilla
& $52.2 \pm 0.40$
& $85.4 \pm 0.00$
& $47.9 \pm 3.37$
& $42.7 \pm 5.32$ \\
w/ \ourmethod
& $\mathbf{67.0} \pm 0.42$
& $\mathbf{87.4} \pm 0.31$
& $\mathbf{57.6} \pm 0.90$
& $\mathbf{74.4} \pm 3.52$ \\
\bottomrule
\end{tabular}
\vspace{4pt}
\caption{Mean performance and standard deviation over three random seeds.}
\label{tab:variance}
\end{table}

\subsection{LLM-as-a-Judge Evaluation of Memory Quality}
\label{appendix:memory_quality_judge}
We further compare \ourmethod{} against Self-Memory using an LLM-as-a-judge evaluation to assess the quality of the constructed memory. For each problem, we provide GPT-5-mini with the task question and two candidate memory entries, one produced by \ourmethod{} and the other produced by Self-Memory. To reduce position bias, we randomize the order of the two candidate memories before presenting them to the judge. The judge is asked to select the memory that is more helpful for solving the task under a unified rubric covering relevance, correctness, actionability, generalization, and safety. Aggregated over the evaluation set, \ourmethod{} achieves a winner rate of 0.72, compared with 0.24 for Self-Memory, with 0.04 ties. This suggests that the memory distilled by \ourmethod{} is judged to be more useful overall, providing more relevant and actionable guidance while maintaining correctness and generality.

\subsection{Additional Case Studies and Examples}
\label{appendix:example}
\label{appendix:full_traj}

This section provides additional examples and details for analyzing the memory constructed by \ourmethod{}. Table~\ref{example} presents representative retrieved trajectory pairs in which agents with different reasoning capacities preserve the same underlying reasoning invariants despite variations in symbolic form, execution order, and implementation detail. These examples show that retrieval can match problems at the level of problem decomposition rather than surface-level reasoning steps.

\begin{figure}[h]
    \centering
    \includegraphics[width=0.8\textwidth]{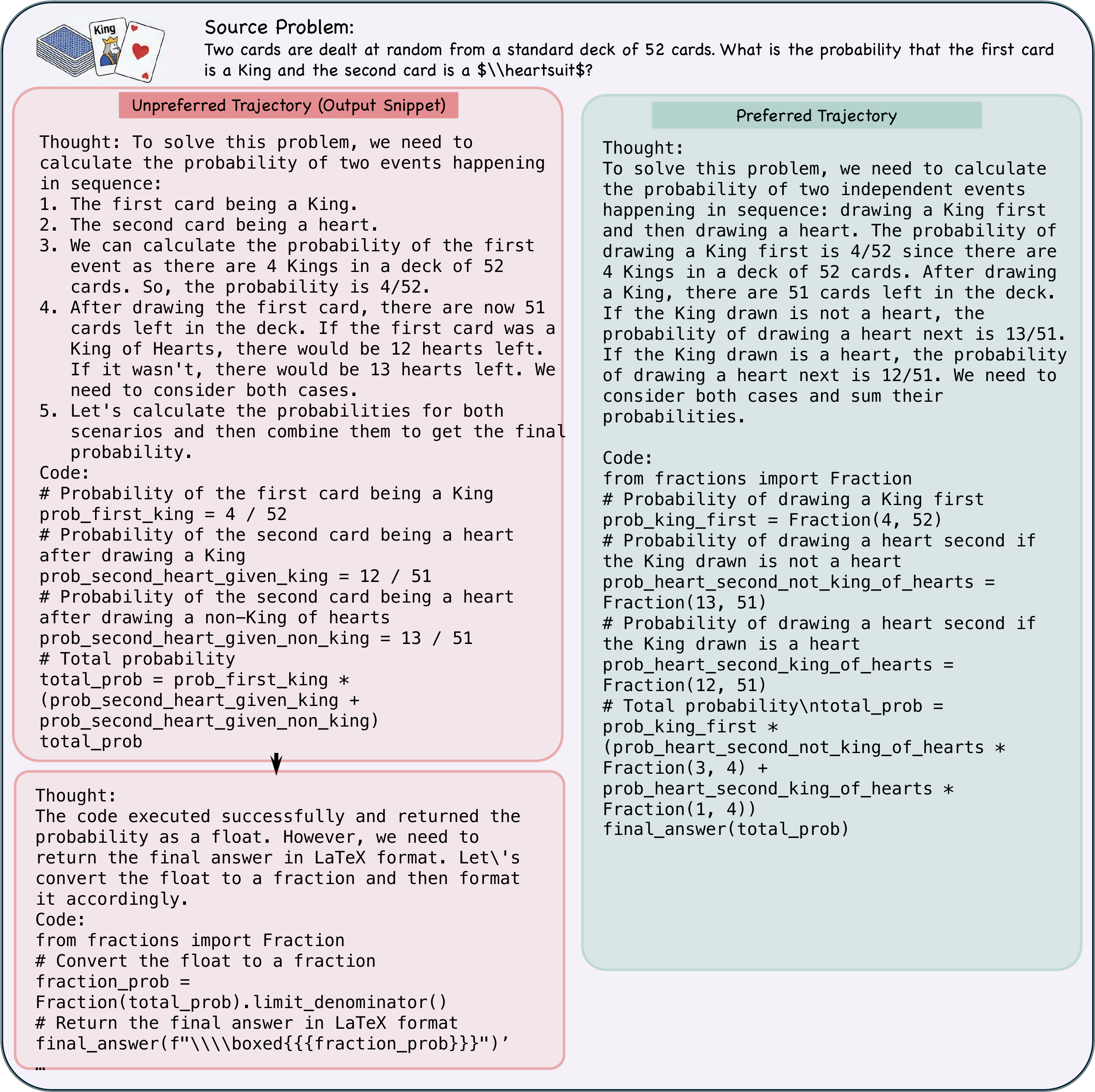}
    \caption{Full trajectory used for memory summarization}
    \vspace{-10pt}
    \label{fig:case_study_full1}
\end{figure}

\begin{figure}[t]
    \centering
    \includegraphics[width=1\textwidth]{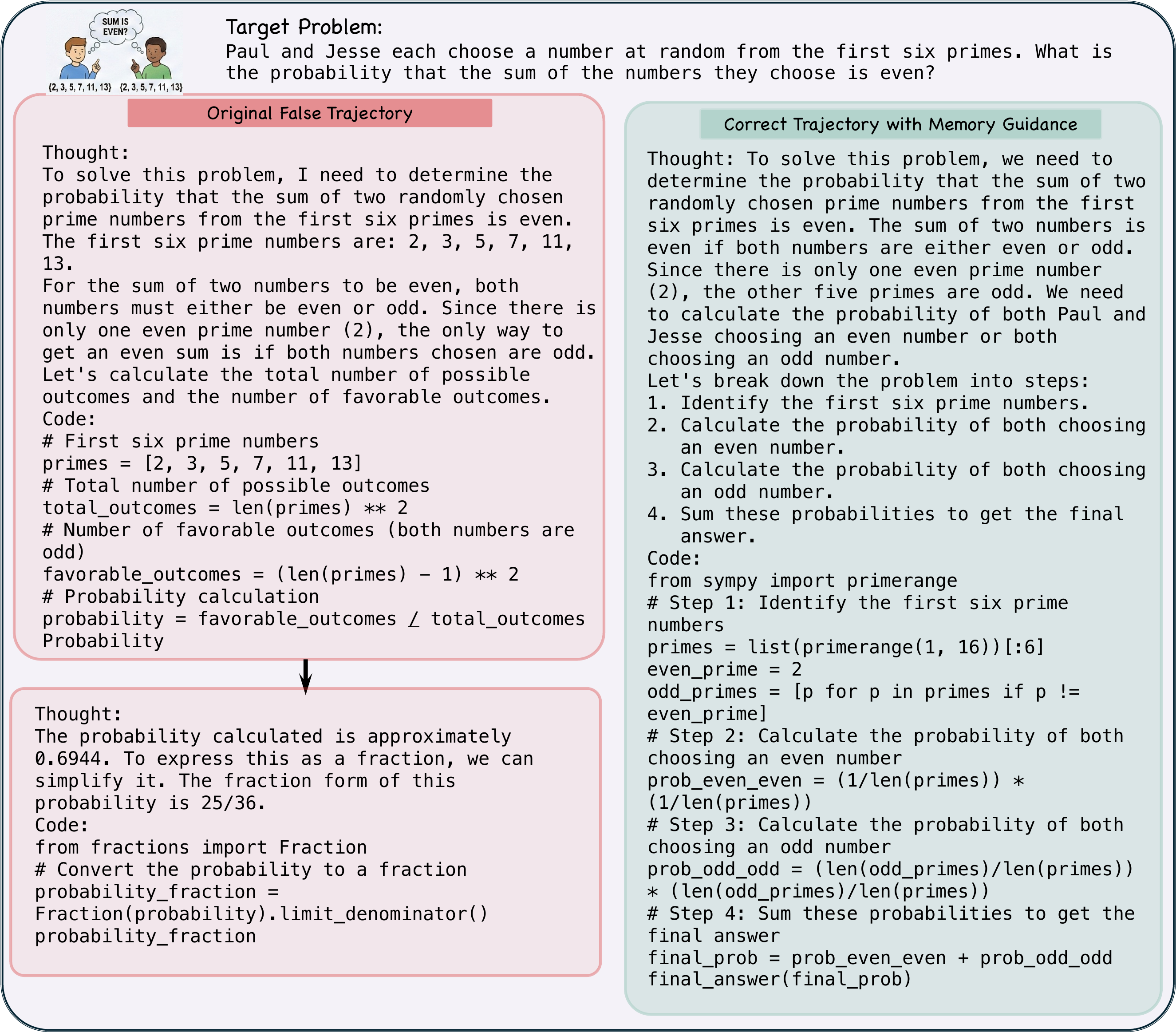}
    \caption{Full trajectory of the memory-augmented inference}
    \vspace{-10pt}
    \label{fig:case_study_full2}
\end{figure}

We also provide the complete trajectories for the case study, including both the memory summarization process and the memory-augmented retrieval in Figure~\ref{fig:case_study_full1} and Figure~\ref{fig:case_study_full2}. The memory used for augmentation is shown in Figure~\ref{fig:case_study}.

\section{Additional Details of \ourmethod}
\label{appendix:retrieval_algorithm}
\paragraph{Shared Memory Clarification.}
We distinguish between a shared memory bank and model-agnostic memory.  A shared memory bank refers to a unified repository constructed from trajectories of multiple heterogeneous agents and accessible to all participating agents.  It does not require every memory entry to be equally useful for every agent. In contrast, fully model-agnostic memory assumes that each memory entry is universally transferable across agents, which may be unrealistic because different backbone models exhibit different failure modes. \ourmethod therefore adopts a shared-bank, model-aware-retrieval design: memory entries are stored in a common bank but annotated with model-identity labels that indicate which agent's failure mode the entry was distilled to address. At inference time, the target agent retrieves entries from the shared bank that match both the task category and its model identity, reducing cross-agent interference while preserving the benefit of collaborative memory construction.

\paragraph{Model- and Task-Aware Memory Retrieval}Algorithm~\ref{alg:retrieve} describes the inference-time retrieval procedure. Given a query and a target agent, \ourmethod first filters memory entries by both the predicted task category and the target model identity, and then ranks the filtered entries by relevance.
\begin{algorithm}[h]
\caption{Model- and Task-Aware Memory Retrieval and Inference}
\label{alg:retrieve}
\begin{algorithmic}[1]
\Require Query $q$; labeled memory bank $\mathcal{M}$; retrieval count $p$; target agent $A_j\in\mathcal{A}$; task classifier $f$
\Ensure Solution $y$
\vspace{0.25em}

\Procedure{Solve}{$q,\mathcal{M},p,A_j$}
    \State $(c_q,u_q) \leftarrow f(q)$ \Comment{LLM-based task classification}
    \State $\ell_j \leftarrow \mathrm{ID}(A_j)$ \Comment{target model identity}
    \State $\mathcal{M}_{q,j} \leftarrow
    \text{Top-}p\!\left(
    \{m\in\mathcal{M}\mid c_m=c_q,\ u_m=u_q,\ \ell_m=\ell_j\}
    \right)$
    \Comment{task- and model-aware retrieval}
    \State $y \leftarrow A_j(q,\mathcal{M}_{q,j})$
    \State \Return $y$
\EndProcedure
\end{algorithmic}
\end{algorithm}

\paragraph{Prompts}
\label{appendix:prompt}
We use the following prompt in Table ~\ref{prompt:single_model}, adapted from Buffer of Thoughts~\citep{yang2024buffer}, to extract high-level reasoning behavior from a model’s single trajectory. We also provide the prompt for contrasting trajectories on the math problems in Table~\ref{prompt:contrast}.

\begin{table*}[tb]
\centering
\small
\begin{tabular}{p{\dimexpr\linewidth-2\tabcolsep}}
\toprule
\begin{minipage}[t]{\linewidth}
\ttfamily

\textbf{Source Question}: \\
Determine the range of parameters for which a triangle exists.\\
\textbf{Summarized Un-preferred Trajectory}: \\
Explicitly writes triangle inequalities and solves the feasible range.\\
\textbf{Summarized Preferred Trajectory}: \\
Uses the same inequality-based feasibility reasoning before computation.\\
\textbf{Memory Guidance}:\\
When determining geometric feasibility, enforce triangle inequalities by converting them into explicit inequality constraints before solving.\\
\textbf{Invariants}:\\
Translate geometric existence into inequality constraints.\\
\textbf{Variants}: \\
One sets equations first, then simplifies; the other simplifies before imposing constraints.
\medskip
\hrule
\medskip

\textbf{Source Question}: \\
How many intersection points are formed by a given number of circles?\\
\textbf{Summarized Un-preferred Trajectory}: \\
Identifies all diagonal pairs and counts only those that intersect, rather than enumerating points directly.\\
\textbf{Summarized Preferred Trajectory}: \\
Counts all circle pairs using $\binom{n}{2}$ and multiplies by the maximum possible intersections per pair.\\
\textbf{Memory Guidance}: 
When dealing with combinatorial geometry problems involving intersections, enforce counting principles by first identifying all possible pairs of elements and then filtering based on intersection criteria; avoid direct enumeration without structured counting.\\
When determining the probability of geometric events, enforce the calculation of favorable outcomes relative to total possible outcomes; avoid assuming uniform distribution without justification.\\
\textbf{Invariants}: Decompose intersection problems into pairwise interactions with bounded intersection counts.\\
\textbf{Variants}: One reasons at the geometric-object level (explicit intersection checks), while the other operates directly at the combinatorial level. \\
One performs conditional filtering; the other applies a closed-form counting bound.

\medskip
\hrule
\medskip

\textbf{Source Question}:Determine a vector onto which multiple vectors have the same projection.\\
\textbf{Summarized Un-preferred Trajectory}:  Sets up dot-product equations equating scalar projections.\\
\textbf{Summarized Preferred Trajectory}:\\
Uses the same projection formula and explicitly enforces equality across vectors.\\
\textbf{Memory Guidance}:\\
When projecting multiple vectors onto the same vector and obtaining identical projections, enforce equality of scalar projection coefficients rather than treating each projection independently.\\
\textbf{Invariants}: \\
Equal projections imply a shared scalar projection constraint.\\
\textbf{Variants}: One sets equations first, then simplifies; the other simplifies before imposing constraints.

\end{minipage} \\
\bottomrule
\end{tabular}
\caption{Illustration of invariant reasoning patterns captured in the summarized memory entries.}
\label{example}
\end{table*}

\begin{table*}[h]
\centering
\small
\begin{tabular}{p{\dimexpr\linewidth-2\tabcolsep}}
\toprule
\begin{minipage}[t]{\linewidth}
\ttfamily
\textbf{Assistant Task}

1. Core task summarization:  
Identify and describe the basic type and core challenges of the problem, such as classifying it as a mathematical problem (e.g., solving a quadratic equation), a data structure problem (e.g., array sorting), or an algorithmic problem (e.g., search algorithms).

2. Solution steps description:  
Outline the general solution steps, including how to define the problem, determine variables, list key equations or constraints, and choose appropriate solving strategies and methods.

3. General answer template:  
Based on the above analysis, propose a reusable template or approach that can be widely applied to this type of problem, including possible variables, functions, or class definitions. For programming problems, provide base classes or interfaces that can be instantiated to solve specific instances. \\
Ensure the response is concise and structured so that instance-specific
solutions can be generalized.

\medskip
\hrule
\medskip

\textbf{Reusable Solution Skeleton}

1. Define symbolic variables or core data structures.  \\
2. Express the main constraints, equations, or problem conditions.  \\
3. Convert the problem into a canonical or solvable form.  \\
4. Apply the appropriate mathematical, algorithmic, or structural method.  \\
5. Solve for the required variable(s) or compute the target quantity.  \\
6. Verify or validate the result using the original constraints or expected outputs.\\
\end{minipage} \\
\bottomrule
\end{tabular}
\caption{Prompt template for single-model reasoning extraction.}
\label{prompt:single_model}
\end{table*}

\begin{table*}[h]
\centering
\small
\begin{tabular}{p{\dimexpr\linewidth-2\tabcolsep}}
\toprule
\begin{minipage}[t]{\linewidth}
\ttfamily

You are an expert analyst for extracting reusable \textbf{REASONING MEMORY}
from contrastive multi-step mathematical reasoning trajectories. You are given multiple incorrect or partially correct reasoning traces
or predicted outputs.\\

Your goal is NOT to solve the problems.  Your goal is NOT to correct the predictions. Your goal is to extract: \\
1) reusable failure-aware reasoning constraints, and \\
2) high-level reasoning strategies that characterize correct multi-step mathematical reasoning, expressed using abstract reasoning rules and high-level mathematical forms (rather than problem-specific calculations).\\

Each extracted strategy must combine:\\
- a trigger (when this strategy should be considered), and  \\
- an enforcement rule (what must be enforced or avoided).\\

\medskip
\textbf{ANALYSIS SCOPE}

Analyze failures and near-misses at the reasoning level, focusing on:\\ 
- explicit formalization of constraints and domains  \\
- correct construction and use of intermediate variables  \\
- alignment between mathematical structure and solution method  \\
- ordering of steps: formalize $\rightarrow$ transform $\rightarrow$ solve $\rightarrow$ verify  \\
- separation between intermediate expressions and final solutions  \\
- handling of extraneous roots, boundary cases, and invalid domains  \\

Use high-level mathematical forms where appropriate, such as:\\
- constraint systems: $\{ f(x)=0,\ g(x)\ge0 \}$  \\
- canonical equations: $ax^2+bx+c=0$, $A\cdot x=b$  \\
- optimization conditions: $\nabla f(x)=0$  \\
- modular forms: $ax\equiv b \pmod n$  \\
- geometric invariants: distance, angle, or area relations  \\

Do NOT focus on:\\
- surface arithmetic mistakes without structural impact  \\
- numerical simplification details  \\
- stylistic presentation or verbosity  \\
- producing corrected solution chains or final answers  \\

\medskip
\textbf{STRICT CONSTRAINTS}

- Each strategy MUST be grounded in an observable pattern across trajectories. \\ 
- Do NOT introduce speculative or unverifiable strategies.  \\
- Strategies MUST be abstract and reusable across different mathematical domains.  \\

\medskip
\textbf{REQUIRED OUTPUT}

Each strategy MUST:\\
- be written as one sentence  \\
- optionally reference high-level mathematical forms or invariants  \\
- follow this format exactly:\\

\texttt{When … , enforce … ; avoid …}\\

At least:\\
- 1 strategy MUST describe failure-prevention or constraint enforcement  \\
- 1 strategy MUST describe high-level reasoning structure or invariants  \\
Do NOT: \\
- include explanations  \\
- reference specific problems, constants, or numeric values  \\
- include worked examples or derivations  \\
- add any text before or after the list  \\

\end{minipage} \\
\bottomrule
\end{tabular}
\caption{Prompt template for contrastive extraction of reusable mathematical reasoning memory.}
\label{prompt:contrast}
\end{table*}
\begin{table*}[t]
\centering
\small
\begin{tabular}{p{\dimexpr\linewidth-2\tabcolsep}}
\toprule
\begin{minipage}[t]{\linewidth}
\ttfamily

You are a mathematical problem classifier. Your task is to assign each math problem to
exactly one subcategory within the given main category.

You will be provided with:\\
1. The problem statement.\\
2. The main category (e.g., Algebra, Precalculus, Geometry, Number Theory).\\
3. A list of allowed subcategories for this category.\\

Your output must be exactly one subcategory from the allowed list.

\medskip
\hrule
\medskip
\textbf{CLASSIFICATION RULES}

1. You MUST choose exactly one subcategory from the provided list.  
   You may NOT output ``None''.  
   You may NOT invent new categories.  
   Only choose from the allowed subcategories provided.

2. If the problem clearly matches a subcategory definition, choose it, even if other
   superficial features appear.

3. If a problem fits multiple subcategories, choose the most specific one.

4. If a problem matches a special enforcement rule, you MUST apply that rule.
   These rules override all ambiguity.

\medskip
\hrule
\medskip
\textbf{SPECIAL ENFORCEMENT RULES}

These rules override ambiguity and enforce consistent labeling:

\begin{itemize}
    \item Polar coordinates, polar--rectangular conversion, parametric coordinates  
    $\rightarrow$ \textbf{Coordinate Systems}.
    
    \item Vector dot products, cross products, projections, vector angles, 3D line geometry  
    $\rightarrow$ \textbf{Analytic Geometry (Vectors \& Lines)}.
    
    \item Distance between points, midpoint, slope, basic coordinate geometry  
    $\rightarrow$ \textbf{Analytic Geometry (Distance \& Midpoints)}.
    
    \item Digit-restricted numbers or divisibility based on digit properties  
    $\rightarrow$ \textbf{Modular Arithmetic \& Digit Properties}.
    
    \item Proper divisors, prime factorization, number of divisors, GCD/LCM  
    $\rightarrow$ \textbf{Divisibility \& Prime Factorization}.
    
    \item Complex number rotation, geometric interpretation, roots of unity, transformations  
    $\rightarrow$ \textbf{Complex Numbers (Geometry \& Transformations)}.
    
    \item Equations involving radicals or exponentials (including logarithms)  
    $\rightarrow$ \textbf{Radical \& Exponential Equations}.  
    \emph{Exception:} Compound interest or exponential growth/decay word problems  
    $\rightarrow$ \textbf{Exponential \& Logarithmic Equations}.
    
    \item Counting or arrangement problems with restrictions (e.g., adjacency constraints)  
    $\rightarrow$ \textbf{Combinatorial Argument \& Constraints}.
    
    \item Constructing polynomials from values or missing coefficients  
    $\rightarrow$ \textbf{Polynomial Interpolation}.
    
    \item Equations involving functional constraints (e.g., $f(x+y)$, invariance)  
    $\rightarrow$ \textbf{Functional Equations}.
\end{itemize}

\medskip
\hrule
\medskip
\textbf{OUTPUT FORMAT}

Your response must be ONLY the name of the chosen subcategory.  
No explanations.  
No justification.  
No additional text.

\medskip
\hrule
\medskip
\textbf{INPUT FORMAT}

Problem:  
\{PROBLEM\}

Main Category:  
\{CATEGORY\}

Allowed Subcategories:  
\{SUBCATEGORY\_LIST\}

\medskip
\textbf{YOUR TASK}

Select exactly one subcategory from the list.
\end{minipage} \\
\bottomrule
\end{tabular}
\caption{Prompt template for task classification used in task-aware retrieval.}
\label{prompt:classifer}
\end{table*}

\section{Limitations.}
\label{limitation}
\ourmethod studies how to construct transferable reasoning memories for heterogeneous agents, but practical deployments may require policy-aware governance over memory access and reuse. For example, different agents, applications, or user groups may be subject to different safety policies or permission constraints. Incorporating access control and safety filtering into collaborative memory retrieval is an important direction for future work.

\end{document}